# Some Strategies to Capture Kāraka-Yogyatā with Special Reference to apādāna

**Swaraja Salaskar[1], Diptesh Kanojia[1,2,3], Malhar Kulkarni[1]**

[1]IIT Bombay

[2]IITB-Monash Research Academy

[3]Monash University

E-mail: swaraja.salaskar77@gmail.com, {diptesh,malhar}@iitb.ac.in

**Abstract**

In today's digital world language technology has gained importance. Several software, have been developed and are available in the field of computational linguistics. Such tools play a crucial role in making classical language texts easily accessible. Some Indian philosophical schools have contributed towards various techniques of verbal cognition to analyze sentence correctly. These theories can be used to build computational tools for word sense disambiguation (WSD). In the absence of WSD, one cannot have proper verbal cognition. These theories considered the concept of 'Yogyatā' (congruity or compatibility) as the indispensable cause of verbal cognition. In this work, we come up with some insights on the basis of these theories to create a tool that will capture Yogyatā of words. We describe the problem of ambiguity in a text and present a method to resolve it computationally with the help of Yogyatā. Here, only two major schools i.e. Nyāya and Vyākaraṇa are considered. Our paper attempts to show the implication of the creation of our tool in this area. Also, our tool involves the creation of an 'ontological tag-set' as well as strategies to mark up the lexicon. The introductory description of ablation is also covered in this paper. Such strategies and some case studies shall form the core of our paper.

**Keywords:** WSD, Yogyatā, Ablation,

## 1. Introduction

Language, is the storehouse of all human knowledge which is represented by words and meanings. Language by itself has an ontological structure, epistemological pinning and grammar. Ambiguity is a feature of natural language. In layman's terms, 'ambiguous' means 'having more than one meaning'. Meanings understood by human beings are based on context, background knowledge, tonal and gestural basis. These factors help them resolve ambiguity to a great extent. With the help of various types of analysis and rationality, human beings can overcome miscommunications caused by ambiguities. There are mainly three types of ambiguities *i.e.,* structural, lexical and semantic. If ambiguity is present in a single word, it is known as lexical ambiguity. Semantic ambiguity means the presence of multiple meanings for the same word. Structural ambiguity, on the other hand, is the presence of two or more possible structures within one single sentence. These types of ambiguities are present in Sanskrit sentences as well. For *e.g.,*

(i) Śītaṁ ghaṭaṁ spṛśati.

(ii) Yānaṁ vanaṁ gacchati.

Possible morphological analysis of these sentences are:

**śītaṁ**
- Gender = n, case = 1$^{st}$, number = singular.
- Gender = n, case = 2$^{nd}$, number = singular.

**ghaṭaṁ**
- Gender = n, case = 1$^{st}$, number = singular.
- Gender = n, case = 2$^{nd}$, number = singular.

**spṛśati**
- dhātu spṛś, person = 3$^{rd}$, Tense = present, number = singular.

The above morphological analysis leads to following possible semantic analyses:

1. Śītam is kartā of an action indicated by spṛś.
2. Śītam is karma of an action indicated by spṛś.
3. Ghaṭam is kartā of an action indicated by spṛś.
4. Ghaṭam is karma of an action indicated by spṛś.
5. Śītam is kartā and ghaṭam is viśeṣaṇa (Adjective) of śītam.
6. Ghaṭam is kartā and śītam is viśeṣaṇa (Adjective) of ghaṭam.
7. Śītam is karma and ghaṭam is viśeṣaṇa (Adjective) of śītam.
8. Ghaṭam is karma; śītam is viśeṣaṇa (Adjective) of ghaṭam

Similarly, also in the second sentence, *i.e.,*

*yānaṁ vanaṁ gacchati*, following morphological analyses are possible:

**Yānaṁ**

- Gender = neuter, case - 1 st, number= singular.
- Gender = neuter, case -2 nd, number= singular.

**vanaṁ**

- Gender = n. case= 1st, number= singular.
- Gender = n. case=2nd, number= singular

**Gacchati**

- It belongs to (dhātu) verbal root gam, (meaning to go) prathama purūṣa (first person) lakāra-laṭ, (present tense) number– ekavacanam (singular).
- Gacchat - saptamī ekavacanam - (Case-7th, number-singular)

This morphological analysis leads to following possible semantic analyses:

1. yānaṁ is kartā of an action indicated by gam
2. yānaṁ is karma of an action indicated by gam
3. vanaṁ is kartā of an action indicated by gam
4. vanaṁ is karma of an action indicated by gam
5. yānaṁ is kartā and vanaṁ is viśeṣaṇa (Adjective) of yānaṁ
6. vanaṁ is kartā and yānaṁ is viśeṣaṇa (Adjective) of vanaṁ
7. yānaṁ is karma and vanaṁ is viśeṣaṇa (Adjective) of yānaṁ
8. vanaṁ is karma and yānaṁ is viśeṣaṇa (Adjective) of vanaṁ

With the analysis above, we observe that the given sentences can be interpreted in multiple ways. While reading such a text, a human being does not notice these ambiguities. Each sentence is made up of words and in Sanskrit language every word has kāraka role to fulfil the meaning of the sentence. Any single word cannot have more than one kāraka role in the same sentence. In this way each dhātu (root word) does have its own expectancy of various kārakas to complete the meaning of the sentence. In the given examples, humans can easily understand that ghaṭam is karma and śītam is viśeṣaṇa (Adjective) of ghaṭam. Whereas, machine cannot reach at such a conclusion.

Logic used by human brain can be explained in one way as follows. In the sentence śītaṁ ghaṭaṁ spṛśati, the word śītaṁ represents the quality and it needs a substance as its locus. The words śītam and ghaṭaṁ have the same case and same gender so śītam should be an adjective.

Similarly, in the second sentence, vehicle has the capacity (yogyatā) to move while forest does not.

For the same sentences machine can have above shown 8 possible analyses. So, the key questions that we try to answer are:

- How will a machine arrive at such a conclusion?
- Can we provide some solutions which may help to prune out such semantic ambiguities?

With the help of verbal cognition (Śābdabodha) theories dealt by various schools of Indian philosophy, the use of semantic constrains like ākāṅkṣā and yogyatā can be proposed. Those can help form rules to prune out other possible analysis. For one of the above sentences, yānaṁ vanaṁ gacchati, the rules would clearly resolve ambiguity. For *e.g.,* the root gam (to go) has an expectancy (ākāṅkṣā) of some kārakas, like kartā (agent), karma (object), adhikaraṇa (locus) etc. There must be a movable entity which can have the compatibility (yogyatā) to be the agent of the action of going, denoted by the root gam. Here, *forest* is not a movable entity but the *vehicle* is. Hence, out of the both above, only *vehicle* can be the agent and not *forest*; and forest will be the karma. In order to translate the logic mentioned above into a computational methodology, we propose the development of a computational database of yogyatā rules which provides structural constraints and prunes out the other analysis of lexemes.

## 2. Linguistic Construct Around Yogyatā

In this section, we describe in detail the construct which leads to the discovery of karaka-yogyatā for sense disambiguation.

### 2.1 Śābdabodha i.e. Verbal Cognition

Theories of Śābdabodha or verbal cognition try to explain the relation amongst the meaning of the words which constitute a sentence. The construction of an intelligible sentence must conform to four conditions i.e. ākāṅkṣā, yogyatā, sannidhi and tātparya. But, in this work, we only deal with first two.

### 2.2 What is meant by yogyatā?

Yogyatā is understood as semantic congruity, suitability or compatibility. Here, the term yogyatā is being used in the sense of mutual compatibility *i.e.,* fitness of the meanings with respect to related words in a linguistic utterance.

## 2.3 What is meant by 'karaka- yogyatā'?

An ability of nouns to get connected with specific action denoted by verbal root as a special relation will be termed as *'kāraka - yogyatā'. i.e.,* Semantic relatedness. According to Paninian grammar, the verb is the most important component of a sentence; and it has a specific expectancy of kārakas. This is nothing but semantic relatedness.

We aim to use these constructs in the development of a parser for the Sanskrit language. We also aim to use the parser and the linguistic constructs described above for disambiguating between word senses using computational algorithms, which in the field of computational linguistics is popularly known as Word Sense Disambiguation (WSD).

## 3. Related work

In order to get rid of the preconceived notion of the yogyatā in question, Ogawa (1997) proposed a new method. He said that yogyatā is a notion which is originally formed in the framework of kāraka theory. Ramanuja Tatacharya (2006) described a collection of theories of śābdabodha. He describes an assembly view of different sastras (nyāya, mīmāṁsā, vyākaraṇa, vedāṁta etc.) and examines theories and subjects. Kunjunni Raja (1968) discusses Indian theories of meanings of different schools. These schools of Indian philosophy discuss yogyatā as a necessary condition for Verbal cognition. We extend it further not just for the śābdabodha, but also use a database as a solution to some problems, as mentioned above in sentences "(i)" and "(ii)" in Section 1.

## 4. Methodology

We aim to extend the ontological tag-set presented by Nair and Kulkarni (2010) and provide an exhaustive set of ontologies. For *e.g.,* the current ontological tag for yānaṁ is acala-nirjīva, but the proposed ontological tag in context of the root word gam for yānaṁ should be gamana-sādhana. We extend the tag-set by providing more such categories.

### 4.1 Process of marking up the lexicon

The procedure which we come up with, for marking up a lexicon is:

- We choose a root word.
- We look for the expectation for various kārakas of the root word.
- We choose a lexeme from the lexicon.
- We tabulate various senses of the lexeme, and check for karaka yogyatā relation of the senses with the root word.
- We mark the lexeme and its senses with kāraka yogyatā relations and store them in our database.

We mark up the lexicon available to us with kāraka yogyatā relations between:

1. dhātu and Word
2. dhātu and a different sense of the word
3. Prefix- dhātu i.e., changed sense of the resultant dhātu with all senses of a word.

| Prefix | Verb | Expectancy of Kāraka | Dict. entry | Sense | Ontological tags |
|---|---|---|---|---|---|
| Optional if any | Any verbal root | kartā | Any dict. entry. | Sense1 | Tag 1 |
| | | Karma | | Sense2 | Tag2 |
| | | karaṇa | | Sense3 | Tag3 |
| | | sampradāna | | Sense4 | Tag4 |
| | | apādāna | | Sense5 | Tag5 |
| | | adhikaraṇa | | Sense6 | Tag6 |
| | | | | Sense8 | Tag7 |

Table 1: Methodology

## 5. Case study of apādāna

Now, we present the case study of an important kāraka known as apādāna. The point of separation is known as apādāna. Hence, it is the source point and it could also be a stationary point[1]. Panini gives various sutras to handle multiple cases of apādāna. Here, we discuss a number of those aphorisms (1.4.24 -1.4.31). The verbal roots covered in these sutras are shown in Table 2, in a representative manner.

These roots expect apādāna kāraka. But which nouns will be eligible to be connected with these roots as an apādāna? In this paper, Kāraka-yogyatā of words beginning with the letter ka have been studied with reference to the above verbal roots.

---
[1] dhruvamapāye apādānam (aṣṭā.1.4.24.)

| zdhruvamapā ye apādānam | bhītrārthān ām bhayahetuḥ | rakṣaṇārtha kāḥ | jugupsāvirāma....upasaṃkh yānam |
|---|---|---|---|
| agi gatyarthe | ñibhī bhaye | gupu rakṣaṇe | drā kutsāyām |
| añcu gatau | dṛ bhaye | pā rakṣaṇe | garha kutsāyām |
| iṇ gatau | bheṣṛ bhaye | deṅ rakṣaṇe | roḍṛ anādare |
| patlṛ gatau | bhyas bhaye | rakṣa pālane | śīṭa anādare |

Table 2: Verbal roots in 1.4.24 -1.4.31

| Root | Expectancy | Lexeme | Senses | Tag | Kāraka-yogyatā |
|---|---|---|---|---|---|
| 1. añcu | All karaka | kaṃsa | Vessel made up of metal | teja:pṛthvīsa ṃyuktaḥ | karma, samprad āna |
|  |  |  | King of mathura | calasajīvaḥ | kartā, karma, karaṇa |
| 2. ñibhī | kartā, apādānam, adhikaraṇa | kaṃsa | Vessel made up of metal | teja:pṛthvīsa ṃyuktaḥ | apādāna m |
|  |  |  | King of Mathura | calasajīvaḥ | kartā, apādāna m |
| 3. pā | kartā, karma apādānam, | kaṃsa | Vessel made up of metal | teja:pṛthvīsa ṃyuktaḥ | karma |
|  |  |  | King of mathura | calasajīvaḥ | kartā, apādāna m |

Table 3: Sample ākāṅkṣā and yogyata

## 6. Yogyatā Relation Tool

As discussed above, we aim to capture the relationship between two words where a dhātu has a kāraka yogyatā relation with a target word. We develop a tool to manually annotate a Sanskrit dictionary with such rules, and store them separately into a database. Our tool is an online web-interface which simultaneously shows the annotator a list of prefixes, a list of dhātu (one dhātu at a time), a list of yogyatā relations, and a list of words from the Monier Williams Dictionary (one word at a time). The tool requires an annotator who will create rules for a pair of words, one of which is a dhātu which may or may not be perpended with a prefix. We call this resultant word L-word. On the other side, a word from the Monier-Williams dictionary is displayed. We call this the R-word. The rule to be created by an annotator requires them to mark every pair of L-word and R-word with a kāraka yogyatā relation. We have an added functionality of appending comments along with the rule for the annotators to justify the rule, if needed. The changed semantics of the dhātu along with the prefix which results in the formation of L-word can also be submitted along with. They can also manually enter the sandhi of the dhātu and prefix i.e., the final L-word in the space provided. For the annotators ease, we provide a Transliteration API on the interface so that *romanized* typing can be facilitated. The tool also provides the functionality to view the rules created for a particular L-word and R-word pair. This enables the annotator to view the work done previously.

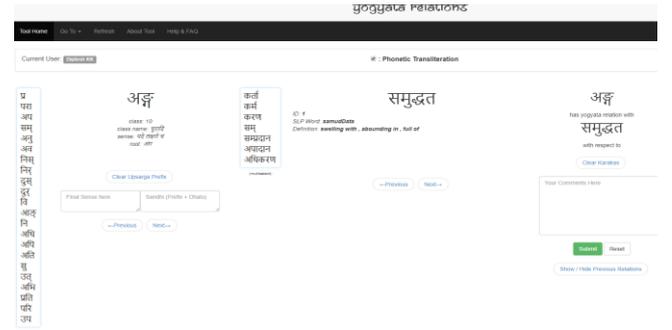

Figure 1: Screenshot of the tool.

### 6.1 System Architecture

Our tool is a web-based interface which is built using PHP on the server side. We use HTML, CSS, and Javascript for front-end creation and manipulating information on the annotator interface. For storage of rules, comments and the annotator data we use MySQL on the backend. Our system uses a PHP-MySQL based login system which requires authentication by the user. Once logged it, a user can easily browse through the L-word and R-word pairs, add rules, view rules and delete them, if needed.

## 7. Output and Results

Following are the some of the outputs:

1. This tool provides yogyata Relation between a dhatu and a lexeme.

2. This tool will give yogyata Relation between a dhatu and various senses of a lexeme.

We also propose the following outputs and functionalities as we form rules for karaka yogyata relations:

1. A view, where, when a lexeme is clicked-all the possible karaka relations with all the dhatu are displayed.

2. A view, where, when a karaka is clicked – all the possible dhatus with which a relation can be formed are displayed.

## 8. Conclusion and Future Work

In this paper, we come up with a methodology for marking lexemes with karaka-yogyatā relations with a dhātu word. We also study the use of ontological tag-sets as a solution for the problem of WSD in NLP, and extend the tag-set previously proposed by others. We develop a tool for marking the Sanskrit lexicon with karaka-yogyatā relations with root words, which stores these relations in a way they can be utilized later for resolving sense disambiguation. Our work proposes to resolve the issue by pruning the number of senses which are available for a lexeme and also via pruning the ontological categories which have the expectancy of a kāraka relation with a root word. We believe that our model, which is currently applicable to the Sanskrit language, should also be applicable to other Indian languages as they borrow a lot of words and are similar in many senses. The methodology in general can be applied to other languages as well.

In future, we would like to analyze and extend the ontological tag-set previously proposed by Nair and Kulkarni (2010) and mark the kāraka yogyatā relations among them. We also aim to annotate more dhatu-word pairs with kāraka yogyatā relations and form a database which can be utilized for solving the problem of WSD and thence for helping NLP applications such as Machine Translation for Sanskrit to other languages and vice versa. We also aim to use Cognitive Psycholinguistics and for verifying if yogyatā is an absolutely necessary condition for verbal cognition. With this, we aim to improve the state of Computational Linguistics for the Sanskrit language with the hope that this impacts other languages as well.

### Acknowledgement

This paper draws insights from an unpublished original Sanskrit work on Yogyatā by Malhar Kulkarni.